\documentclass{article}
\usepackage{amssymb}
\usepackage{amsmath}
\usepackage{graphicx}
\usepackage{stfloats}
\usepackage{float}
\usepackage{booktabs}
\usepackage{wrapfig}
\usepackage{caption}
\usepackage{multirow}
\usepackage{subcaption}
\usepackage{pifont}
\usepackage{enumitem}
\usepackage[table]{xcolor}
\usepackage{hyperref}

\usepackage[preprint]{corl_2025} 

\definecolor{purple}{rgb}{0.65,0,0.65}

\title{VISTA: Generative Visual Imagination for Vision-and-Language Navigation}

\author{
  Yanjia Huang \quad
  Mingyang Wu \quad
  Renjie Li \quad
  Zhengzhong Tu\footnotemark \\
  Department of Computer Science and Engineering, Texas A\&M University \\
  \href{https://taco-group.github.io/}{\texttt{https://taco-group.github.io/}}
}

\begin{document}
\maketitle
\renewcommand{\thefootnote}{\fnsymbol{footnote}}  
\footnotetext{Corresponding author: \href{mailto:tzz@tamu.edu}{tzz@tamu.edu}}



\begin{abstract}  
Vision-and-Language Navigation (VLN) tasks agents with locating specific objects in unseen environments using natural language instructions and visual cues. 
Many existing VLN approaches typically follow an `observe-and-reason' schema, that is, agents observe the environment and decide on the next action to take based on the visual observations of their surroundings.
They often face challenges in long-horizon scenarios due to limitations in immediate observation and vision-language modality gaps.
To overcome this, we present VISTA, a novel framework that employs an `imagine-and-align navigation strategy. 
Specifically, we leverage the generative prior of pre-trained diffusion models for dynamic visual imagination conditioned on both local observations and high-level language instructions. A Perceptual Alignment Filter module then grounds these goal imaginations against current observations, guiding an interpretable and structured reasoning process for action selection.
Experiments show that VISTA sets new state-of-the-art results on Room-to-Room (R2R) and RoboTHOR benchmarks, \textit{e.g.}, +3.6\% increase in Success Rate on R2R.
Extensive ablation analysis underscores the value of integrating forward-looking imagination, perceptual alignment, and structured reasoning for robust navigation in long-horizon environments.
\end{abstract}



\section{Introduction}
	
   \textit{“Imagination is more important than knowledge. For knowledge is limited, whereas imagination embraces the entire world.”}  
\hfill — Albert Einstein

   Humans rarely navigate by sight alone. When sitting in sofa instructed by your friend to "go to kitchen and grab a soda", we don't only scan our surroundings; instead we imagine what the kitchen looks like, anticipate its layout, and mentally simulate possible paths, even we are in a friend's house. This cognitive loop between imagination, perception, and reasoning enables us to act with foresight, adapt to ambiguity, and recover from uncertainty. Despite rapid progress in Vision-and-Language Navigation (VLN), most existing agents operate without such a mechanism \cite{anderson2018visionandlanguagenavigationinterpretingvisuallygrounded, wang2019reinforcedcrossmodalmatchingselfsupervised, wu2024voronavvoronoibasedzeroshotobject}.  They treat navigation as a sequence of local predictions, reacting to immediate observations and static instructions without explicitly modeling what lies beyond their current view \cite{pengying2024voronav, samir2022cows, congcong2024zeroshot, yinpei2023think}.
   

  
In this paper, we argue that explicit goal imagination is crucial for embodied navigation, enabling them to anticipate future states and make more informed decisions. Integrating visual imagination into navigation presents significant challenges \cite{xin2018reinforced, jinyu2022reinforced, gengze2023navgpt, vishnu2022clipnav, yuxing2023discuss, liuyi2024causalitybased, alex2021episodic, liuyi2023pasts, jialu2023improving}.  One key difficulty is to determine when to rely on high-level semantic instructions, especially under trajectory uncertainty. Moreover, aligning imagined goals with current observations requires robust and interpretable.

To address these challenges, we propose \textbf{VISTA}—a cognitively inspired VLN system equipping the agent to imagine, verify, and act through a structured, interpretable framework. Specifically, we introduce the Adaptive Imagination Scheduler (AIS), a mechanism that dynamically decides between instruction-driven and observation-driven imagination modes based on trajectory confidence and visual-semantic alignment. Crucially, rather than treating these generated visuals as uninterpretable priors \cite{ko2023learningactactionlessvideos, chen2022think, chen2021history}, we apply a Perceptual Alignment Filter (PAF) module that aligns imagined goals with current observations via cross view attention. This yields interpretable visual attention maps that highlight goal-relevant regions — effectively letting the agent see where it should look before deciding where to go. Finally, all information is passed to a navigational reasoning module that performs Chain-of-Thought (CoT) \cite{wei2023chainofthoughtpromptingelicitsreasoning} action prediction, closing the loop with language, vision, and decision-making. The AIS and PAF components together form a robust solution, effectively enabling forward looking imagination to improve navigation decisions.

Our contributions are summarized as follows:
\begin{itemize}[leftmargin=*]
\item 
We present a unified closed-loop VLN framework—VISTA, integrating proactive imagination, visual-semantic alignment, and structured Chain-of-Thought reasoning, resulting in robust and interpretable navigation behaviors. 
\item We introduce two modules: Adaptive Imagination Scheduler (AIS), dynamically selecting between instruction-driven and observation-driven imagination modes based on trajectory uncertainty and visual-semantic alignment; and Perceptual Alignment Filter (PAF) module, explicitly grounding visual imagination with real-time observations to enhance interpretability and navigation accuracy. 
\item We validate our approach on standard VLN benchmarks ( R2R \cite{anderson2018visionandlanguagenavigationinterpretingvisuallygrounded}, 
 RoboTHOR \cite{deitke2020robothoropensimulationtorealembodied}), demonstrating significant gains in success rate, navigation efficiency, and interpretability.
\end{itemize}

\section{Related Work}
\subsection{Vision-and-Language Navigation (VLN)}

Vision-and-Language Navigation (VLN) aims to guide agents through unfamiliar environments using natural language instructions. This task has been enabled by large-scale simulators (e.g., Matterport3D \cite{chang2017matterport3dlearningrgbddata}) and benchmark datasets such as R2R \cite{anderson2018visionandlanguagenavigationinterpretingvisuallygrounded} and REVERIE \cite{qi2020reverieremoteembodiedvisual}, which focus on low-level navigation and high-level object interaction respectively. To further enhance generalization to unseen environments, recent approaches leverage large-scale pretraining with vision-language objectives \cite{tan2019learning, liu2021vision, fu2020counterfactual, congcong2024zeroshot}.

Recent works have explored incorporating large language models (LLMs) as navigation backbones \cite{zhou2023navgptexplicitreasoningvisionandlanguage, chen2024mapgptmapguidedpromptingadaptive, long2023discussmovingvisuallanguage} to enhance VLN agents. For instance, NavGPT\cite{gengze2023navgpt} treats navigation as a language-driven process, converting observations and history into prompts for LLM interpreting and reasoning. DiscussNav leverages dialogue-style interactions with LLM experts to enable decision-making, while MapGPT \cite{chen2024mapgptmapguidedpromptingadaptive} introduces an online-updated, language-formalized map to support long-range planning and adaptive behaviors. NaviLLM \cite{zheng2024learninggeneralistmodelembodied} further explores fine-tuning LLMs for embodied tasks. Despite these advances, current VLN agents still struggle to adapt to realistic, commonsense rich environments and often lack interpretability and structured reasoning in decision-making. Unlike these methods, we integrates visual imagination and perceptual alignment into structured reasoning enabling more interpretable and goal driven navigation decisions.

   \begin{figure*}[htbp!]
    \centering
    \includegraphics[width=1\linewidth]{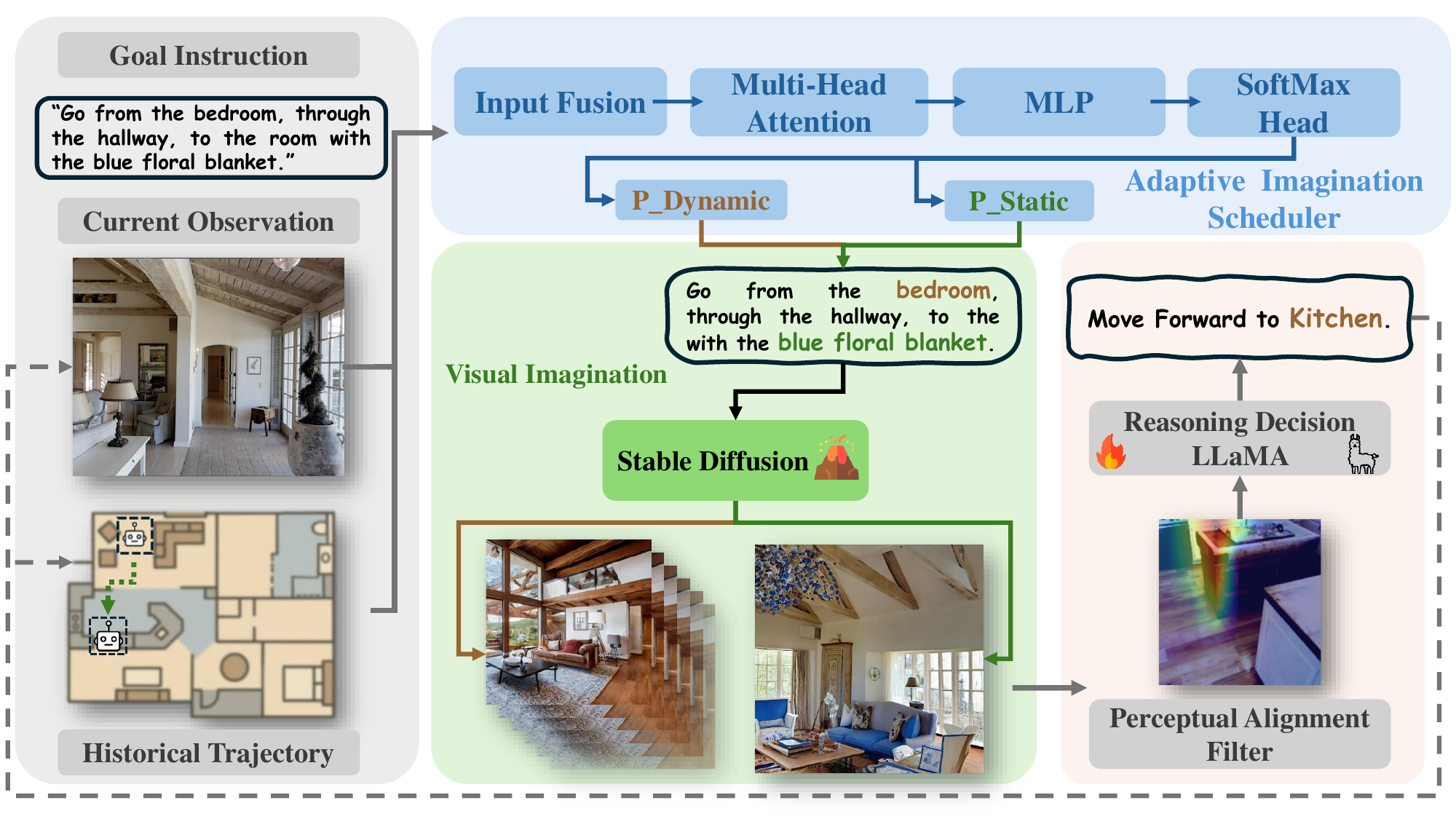}
    \caption{Overview of \textbf{VISTA}---At each step, the agent predicts what it expects to see, imagines that goal visually, aligns the prediction with reality, and reasons through language to decide how to move. 
This loop is guided by an adaptive scheduler that balances global task intent with local observations, enabling dynamic goal prediction. 
By grounding imagination in perception and embedding structured reasoning into decision-making, VISTA offers a forward-looking and interpretable approach to vision-and-language navigation.
}
    \label{fig:overview}
\end{figure*}

\subsection{Visual Imagination in VLN}
Recent research has explored visual imagination through generative models to enhance VLN agents by predicting future observations. Pathdreamer~\cite{koh2021pathdreamerworldmodelindoor} uses GANs to hallucinate panoramic views from visual history, but its outputs are constrained to heuristic-based search and lack policy integration. PanoGen~\cite{li2023panogentextconditionedpanoramicenvironment} and DiffPano~\cite{ye2024diffpanoscalableconsistenttext} utilize diffusion models to extrapolate goal views. PanoGen++~\cite{Wang_2025} further adapts diffusion models for indoor navigation via LoRA fine-tuning and structured inpainting. However, these approaches typically rely on offline generation and treat imagined content as auxiliary information rather than core component of decision-making.

\subsection{Chain of Thoughts Reasoning}
Building upon the integration of LLMs into VLN agents, recent research has further emphasized  the need for advanced reasoning strategies to strengthen decision-making. Among them, Chain-of-Thought (CoT) prompting \cite{wei2022chain_of_thought} introduces an in-context learning approach that decompose complex reasoning tasks into a sequence of simpler, interpretable sub-steps. Extending the linear reasoning paradigm, Tree-of-Thoughts (ToT) \cite{yao2023tree_of_throught} adopts a branching structure that enables the parallel exploration of multiple reasoning paths. Self-consistency \cite{wang2022self_consistency} further augments CoT by sampling diverse reasoning chains and aggregating the outputs through majority voting. Inspired by reasoning strategies, C-Instructor \cite{kong2024controllable_instructor} employs CoT prompting to guide LLMs in identifying key landmarks and generating navigation instructions. InstructNav \cite{long2024instructnav} introduces a Dynamic Chain-of-Navigation, aligned with CoT reasoning, to unify diverse instruction types and convert linguistic planning into executable navigation trajectories. NavCoT~\cite{lin2025navcot} further introduces a trainable CoT framework that incorporates world model reasoning and formalized supervision. Our method, VISTA, rather than using LLMs as monolithic planners, we embeds visual imagination and alignment within the CoT framework, creating visually grounded reasoning steps.

\section{Method}
    We propose a cognitively inspired navigation framework operating in a closed-loop reasoning cycle, where the agent actively imagines future goals, aligns these predictions with real-time visual input, and reasons over actions. Specifically, at each navigation step, an Adaptive Imagination Scheduler selects modes based on two metrics. The chosen mode generates a visual hypothesis of the next goal through Visual Imagination Module , which is then spatially aligned with current observations via Perceptual Alignment Filter. Finally, the agent integrates this aligned visual information with instructions and navigation history through Navigational Chain-of-Thought Reasoning to select next action. This structured imagination to alignment to reasoning loop enables the agent to proactively anticipate goals, ground predictions visually, and make interpretable navigation decisions, shown in Figure\ref{fig:overview}.

\subsection{Adaptive Imagination Scheduler}

\begin{figure*}
    \centering
    \includegraphics[width=1\linewidth]{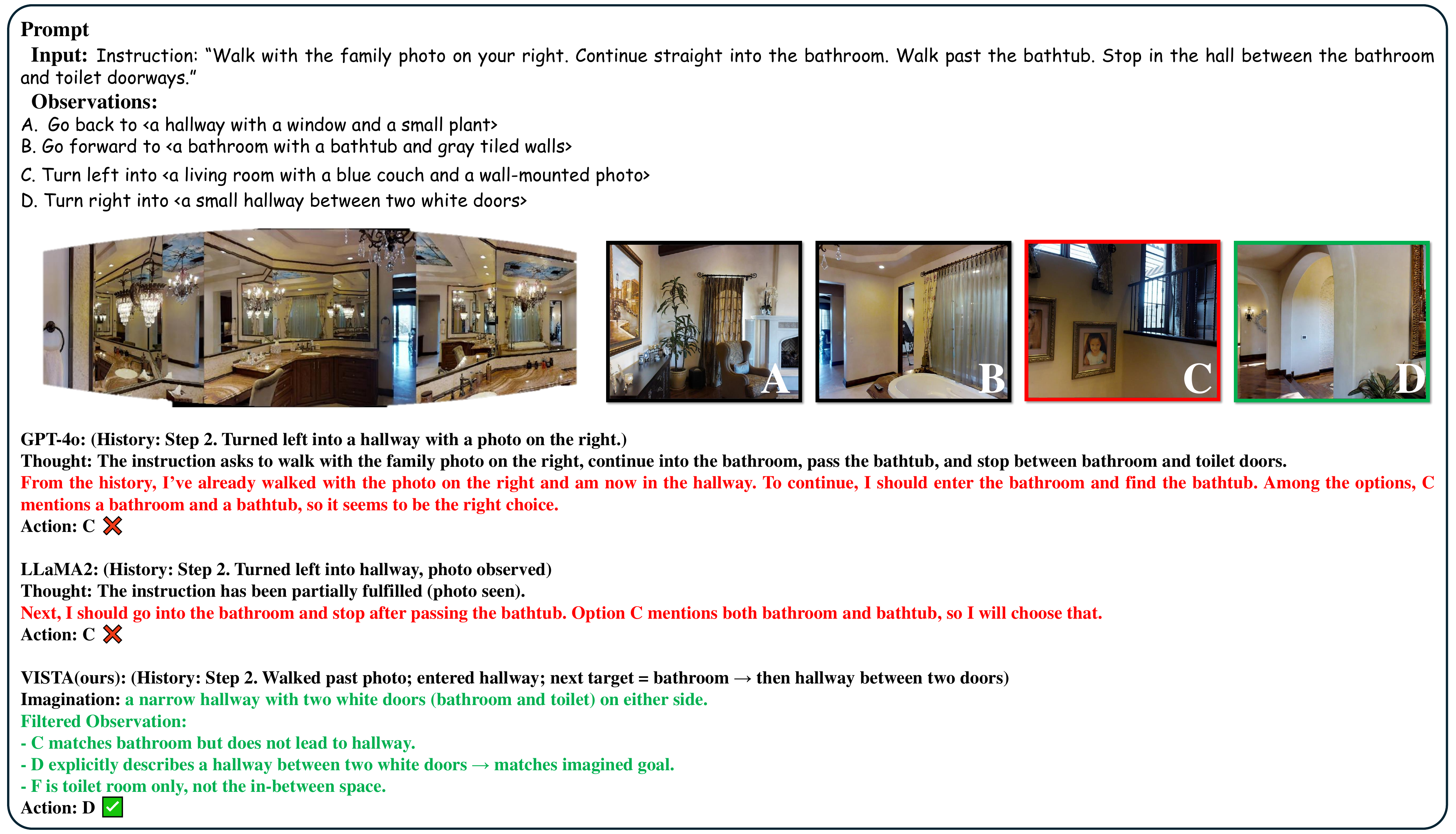}
    \caption{Chain-of-Thought (CoT) reasoning comparison on a navigation step. 
Given the same instruction, observation candidates, and history, we compare zero-shot GPT-4, LLaMA2 baseline, and our VISTA agent.
While other methods rely on textual cues only, our agent grounds its reasoning in visual imagination and attention alignment.
VISTA successfully identifies the correct observation (D) by matching imagined scene with actual perception.}
    \label{fig:Prompt}
\end{figure*}

A key challenge in goal anticipation is deciding when an agent should rely on high-level instructions and when to adapt based on immediate visual observations. Instructions provide semantic guidance but can lack precision, especially in unfamiliar settings, where visual observations are detailed but potentially "noisy" without context. To balance these two sources effectively, we propose an \textbf{Adaptive Imagination Scheduler} \textbf{(AIS)}. It dynamically switches between \textbf{static }(instruction driven) and \textbf{dynamic} (observation driven) goal predictions.

Specifically, in static mode, AIS extracts semantic entities (e.g., "kitchen", "stairs", "sofa") from instructions with language models. In dynamic mode, it leverages last trajectory history and current visual observations to propose immediate goals. AIS determines the prediction mode by evaluating two metrics: \textit{trajectory uncertainty} $u_t$, measured by action entropy and path deviation. And \textit{visual semantic similarity} score $s_t$ which computed as the cosine similarity between embeddings of imagined and observed images. Formally, we define the mode selection as: 
\begin{equation}
z_t = 
\begin{cases}
\text{static}, & \text{if } u_t < \tau_u \text{ and } s_t > \tau_s \\
\text{dynamic}, & \text{otherwise}
\end{cases}
\end{equation}
where  $\tau_u$ and $\tau_s$ are predefined thresholds. AIS thus enables the agent to flexibly integrate high-level intent with real-time perceptual insights, facilitating robust and context-aware navigation decisions.

\subsection{Visual Imagination Module}
\label{sec:Image Generation}
To visually anticipate future goals, our agent constructs scene hypotheses using a two-stage vision-language pipeline. First, a candidate scene—either from training data or previously imagined—is passed through Qwen2-VL~\cite{Qwen2-VL} to generate a descriptive caption (e.g., \textit{"A spacious living room with a blue sofa, wooden floors, and large windows."}). This caption is then fed into a LoRA-finetuned Stable Diffusion model~\cite{rombach2021highresolution,hu2021loralowrankadaptationlarge} to synthesize a corresponding visual scene. To enhance contextual relevance, we additionally create an inpainted version $I_t^{\text{inpaint}}$ by filling masked regions of the current observation with imagined content guided by the same caption. This approach produces semantically rich, visually coherent hypotheses that serve as proactive priors for subsequent alignment and decision-making steps.

\subsection{Perceptual Alignment Filter}

Given its imagined visual hypothesis, the agent must verify whether its prediction aligns with reality—and if so, where to look. To this end, we introduce a perceptual alignment filter that compares the agent’s imagined goal scenes with its current egocentric observation, producing an attention map that highlights likely goal regions. The filter operates over three visual inputs: the current observation $O_t$, the imagined scene $I_t^{\text{SD}}$, and the inpainted image $I_t^{\text{inpaint}}$. Each image is encoded through a shared ResNet-18 backbone, and their features are fused via a multi-head attention mechanism. The fused representation is decoded into a spatial attention map $A_t \in [0,1]^{H \times W}$ that assigns soft relevance scores to each pixel in $O_t$:

\begin{equation}
A_t = \sigma\left( f_{\theta}\left( \phi(O_t), \phi(I_t^{\text{SD}}), \phi(I_t^{\text{inpaint}}) \right) \right)
\end{equation}

To supervise the attention map, we use pseudo ground-truth masks $A_t^{\text{GT}}$ derived from grounding models. Specifically, we extract entity-level keywords from captions  generated by Qwen, and localize them in $O_t$ using Grounded-SAM \cite{ren2024groundedsamassemblingopenworld}. These masks indicate where the imagined object should appear in the real observation.

We construct a dataset of 30,000 quadruples—$(O_t, I_t^{\text{SD}}, I_t^{\text{inpaint}}, A_t^{\text{GT}})$—spanning diverse scene types and object categories. The filter is trained using a combination of binary cross-entropy and soft Dice loss, with Adam optimizer and a learning rate of $1 \times 10^{-4}$. All images are resized to $256 \times 256$, and the model is trained for 50 epochs with early stopping. This alignment mechanism enables the agent to reconcile imagination with perception transforming generative predictions into spatially grounded attention. Crucially, the resulting mask $A_t$ serves as a visual bottleneck: it filters irrelevant information, enforces goal specific focus, and enhances the interpretability of downstream decisions.

\subsection{Navigational Chain-of-Thought Decision}

After aligning its imagined goal with current observations, the agent must determine the next navigation step. Unlike traditional flat action predictions, we adopt a structured, language-grounded Chain-of-Thought (CoT) reasoning \cite{lin2025navcotboostingllmbasedvisionandlanguage, wei2022chain_of_thought} approach. At each timestep, our policy model explicitly decomposes decision-making into an interpretable, step-wise reasoning process that integrates multiple modalities—including task instructions, observation captions, visual attention maps, and navigation history—into structured prompts. Our CoT reasoning consists of three stages:
\ding{182} Goal grounding: Infer what the agent is currently trying to find (e.g., “a room with a blue sofa”) based on the global instruction.
\ding{183} Perceptual verification: Interpret the observation and attention map to determine whether this goal appears in the current scene.
\ding{184} Decision justification: Based on alignment, select the best next action and justify it.

To generate interpretable reasoning traces, we fine-tune a LLaMA2-7B \cite{touvron2023llama2openfoundation} model with LoRA adapters to output both intermediate reasoning steps and final actions. We represent visual context succinctly via textual summaries derived from attention maps (e.g., "the mask highlights the left side of the hallway"). Crucially, unlike previous CoT approaches that primarily served post-hoc explanations, our CoT reasoning is directly embedded into the navigation control loop, bridging abstract instructions with concrete visual perception. This explicit, step-wise reasoning enhances interpretability, improves navigation robustness, and supports more transparent decision-making processes. A complete CoT reasoning example is illustrated in Figure~\ref{fig:Prompt}.

 \begin{table*}[t]
\centering
\small
\renewcommand{\arraystretch}{1.1}
\setlength{\tabcolsep}{6pt}
\begin{tabular}{lcccccccc}
\toprule
\multirow{2}{*}{\textbf{Models}} & \multicolumn{4}{c}{\textbf{Validation Unseen}} & \multicolumn{4}{c}{\textbf{Test Unseen}} \\
\cmidrule(lr){2-5} \cmidrule(lr){6-9}
& TL $\downarrow$ & NE $\downarrow$ & SR $\uparrow$ & SPL $\uparrow$ 
& TL $\downarrow$ & NE $\downarrow$ & SR $\uparrow$ & SPL $\uparrow$ \\
\midrule
EnvDrop ~\cite{tan2019learning}& 10.70 & 5.22 & 52.0 & 48.0 & \textbf{11.66} & 5.23 & 51.0 & 47.0 \\
PREVALENT~\cite{hao2020towards}       & \textbf{10.19} & 4.71 & 58.0 & 53.0 & \textbf{10.51} & 5.30 & 54.0 & 51.0 \\
RecBERT~\cite{hong2020recurrent}          & 12.01 & 3.93 & 63.0 & 57.0 & 12.35 & 4.09 & 63.0 & 57.0 \\
HAMT~\cite{chen2021history}           & 11.46 & 3.62 & 66.2 & 61.5 & 12.27 & 3.93 & 65.0 & 60.0 \\
HAMT-Imagine~\cite{perincherry2025visualimaginationsimprovevisionandlanguage}& 11.80 & \textbf{3.58} & \textbf{67.26} & \textbf{62.02} & 12.66 & \textbf{3.89} & \textbf{65.0} & \textbf{60.0} \\
DUET~\cite{chen2022think}               & 13.94 & 3.31 & 71.52 & 60.41 & 14.73 & 3.65 & 69.0 & 59.0 \\
DUET-Imagine~\cite{perincherry2025visualimaginationsimprovevisionandlanguage}& 14.35 & \textbf{3.19} & \textbf{72.12} & \textbf{60.48} & 15.35 & \textbf{3.52} & \textbf{71.0} & \textbf{60.0} \\
PanoGen~\cite{li2023panogentextconditionedpanoramicenvironment}       & 13.40 & 3.03 & 74.2 & 64.3 & 14.38 & \textbf{3.31} & \textbf{71.7} & \textbf{61.9} \\
 NavCoT~\cite{lin2025navcotboostingllmbasedvisionandlanguage} & \textbf{9.95}& 6.26& 40.23& 36.64& /& /& /&/\\
\bottomrule
 \cellcolor{blue!15} VISTA (ours)& \cellcolor{blue!15} 13.26& \cellcolor{blue!15} \textbf{2.92}& \cellcolor{blue!15} \textbf{77.8}& \cellcolor{blue!15} \textbf{68.3}& \cellcolor{blue!15}\textbf{14.20}& \cellcolor{blue!15}\textbf{3.77}& \cellcolor{blue!15}\textbf{74.9}& \cellcolor{blue!15}\textbf{66.7}\\
 \bottomrule
\end{tabular}
\caption{
Performance comparison on the R2R dataset. VISTA achieves state-of-the-art performance across all core metrics.
Compared to the previous best model DUET-Imagine, our method improves SR by \textbf{+3.2\%} while maintaining a significantly lower navigation error.
}
\label{tab:r2r_results}
\end{table*}

\section{Experiments}
\subsection{Experimental Setup}
We evaluate our approach on the Room-to-Room (R2R) dataset \cite{anderson2018visionandlanguagenavigationinterpretingvisuallygrounded}, built within the Matterport3D simulator \cite{chang2017matterport3dlearningrgbddata}. The environment consists of 90 real-world indoor scenes, with 10,567 panoramic viewpoints connected as a navigation graph. Each panorama is rendered as 36 discretized single-view images, providing a diverse egocentric perspectives at each location. The dataset provides with 7,189 trajectories, each annotated with three natural language instructions and a ground-truth path. It is split into: train (4675 trajectories), val-seen (340 trajectories), val-unseen (783 trajectories), and test (1391 trajectories) \cite{perincherry2025visualimaginationsimprovevisionandlanguage}. To support imagination-guided navigation, we curate a large-scale auxiliary dataset—R2R-Imagine—which provides visual predictions conditioned on both instructions and predicted goal concepts. In total, we generate over \textbf{250,000 }images (256*256), covering imagined goal scenes across all R2R splits. These images are produced via a two-stage vision-language process (see Section~\ref{sec:Image Generation}) Image generation takes approximately 4.2 seconds per image on a single NVIDIA A100 GPU. The full R2R-Imagine dataset—including all instruction-aligned, caption-conditioned imaginations—will be released publicly upon acceptance to encourage further research on generative visual priors for navigation. Visual goal captions are generated using Qwen2-VL, and converted into imagined goal images via LoRA-finetuned Stable Diffusion (v1.5). LoRA is trained using 5k paired goal-caption samples from the MP3D environment, with a learning rate of 5e-5 and 50 epochs. Training takes approximately 18 hours on 2 NVIDIA A100 GPUs with 80GB memory, using a batch size of 32 and sequence length of 30.  


\begin{figure}[t]
  \centering
  \begin{subfigure}[b]{0.60\textwidth}  
    \centering
    \includegraphics[width=\textwidth]{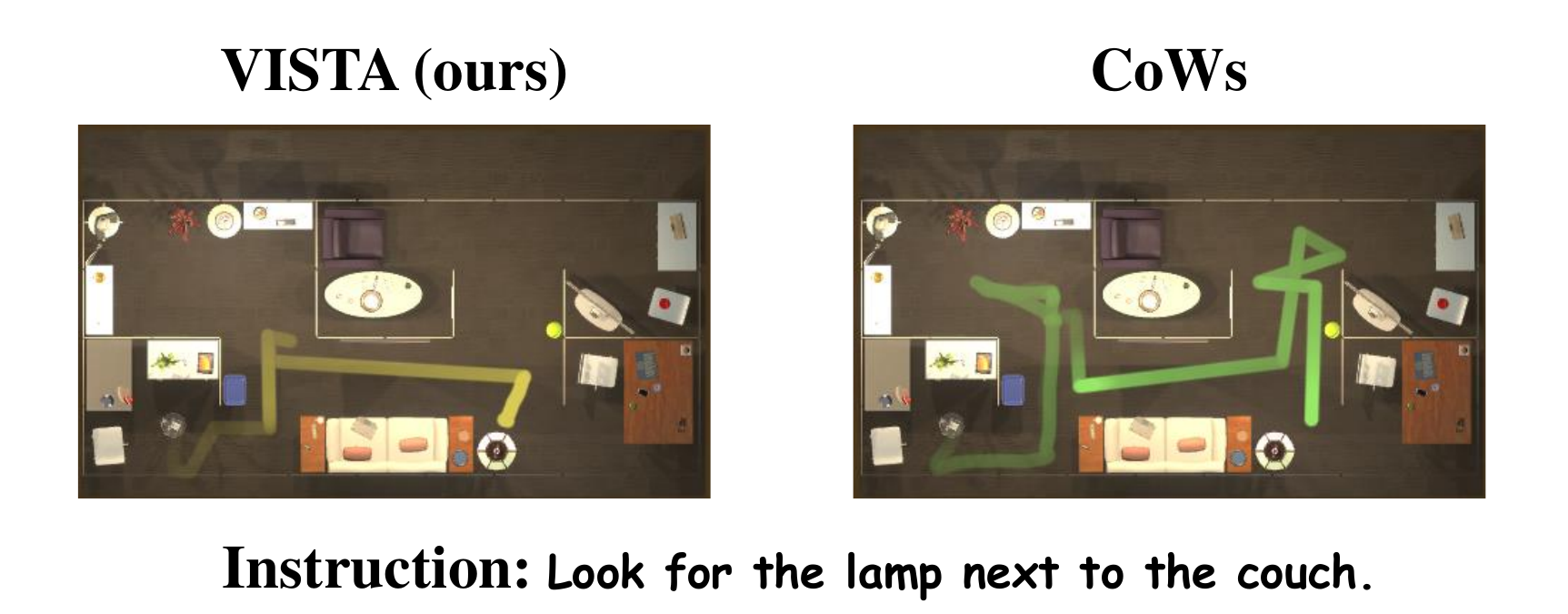}
    \caption{Instruction-following comparison with CoWs.}
    \label{fig:robothor-image}
  \end{subfigure}
  \hfill
  \begin{subfigure}[b]{0.38\textwidth}  
    \centering
    \small
    \begin{tabular}{lcc}
    \toprule
    \textbf{Model} & \textbf{SWPL↑} & \textbf{SR↑} \\
    \midrule
    CLIP-Ref. \cite{samir2022cows}     & 2.4  & 2.7  \\
    CLIP-Patch \cite{samir2022cows}   & 10.6 & 20.3 \\
    CLIP-Grad. \cite{samir2022cows}   & 9.7  & 15.2 \\
    MDETR \cite{samir2022cows}        & 8.4  & 9.3  \\
    OWL \cite{samir2022cows}          & 17.2 & 27.5 \\
    ESC \cite{zhou2023escexplorationsoftcommonsense} & 22.2 & 38.1 \\
    VLTNet \cite{congcong2024zeroshot} & 17.1 & 33.2 \\
    \bottomrule
   \cellcolor{blue!15} \textbf{VISTA (ours)} & \cellcolor{blue!15} \textbf{28.8} & \cellcolor{blue!15} \textbf{43.1} \\
    \end{tabular}
    \caption{Comparison on RoboTHOR (Table).}
    \label{tab:robothor-table}
  \end{subfigure}
  \caption{\textbf{Comparison on RoboTHOR.} (a) Qualitative comparison between VISTA and CoWs. (b) Table comparison of different models.}
  \label{fig:robothor-both}
\end{figure}

\noindent
\textbf{Evaluation Metrics}
We employ standard VLN metrics~\cite{ilharco2019general}: \ding{182} Success Rate (SR) measures the percentage of episodes where the agent stops within 3 meters of the goal; \ding{183} Success weighted by Path Length (SPL) penalizes longer trajectories by comparing the agent’s path to the shortest possible route; \ding{184} Navigation Error (NE) is the shortest-path distance between the agent’s final position and the goal location, measured in meters; \ding{185} Trajectory Length (TL) is the total distance traveled by the agent during an episode. Higher SR and SPL indicate better performance, while lower NE and TL reflect more accurate and efficient navigation. During inference, VISTA runs at an average speed of 8 FPS on an NVIDIA A100 GPU, including visual imagination and CoT reasoning, demonstrating practical feasibility.


 \subsection{Main Results}

\begin{figure*}[t]
  \centering
  \begin{subfigure}[t]{0.45\textwidth}
    \centering
    \includegraphics[width=\textwidth]{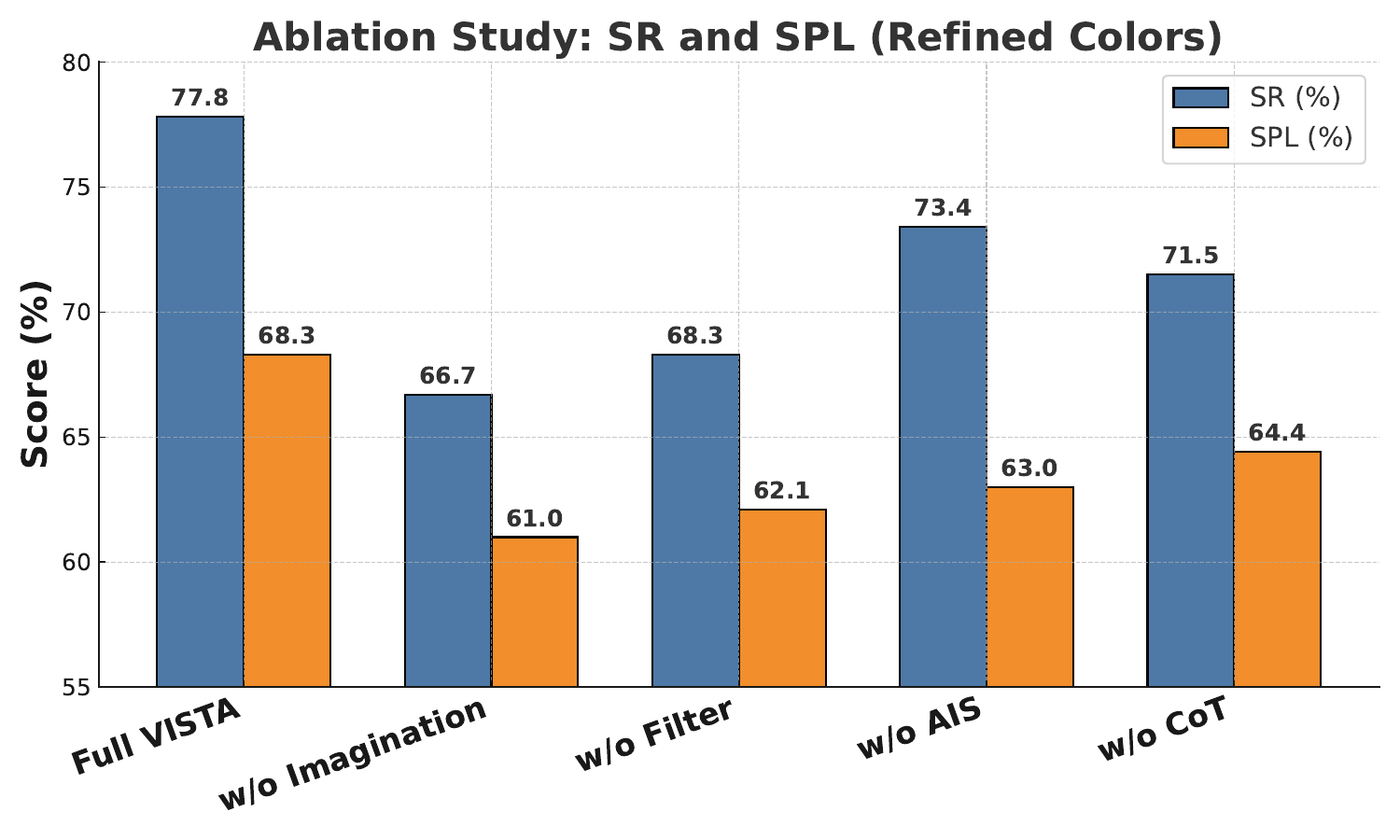}
    \caption{Removing imagination or alignment modules significantly drops navigation success.}
    \label{fig:ablation-bar}
  \end{subfigure}
  \hfill
  \begin{subfigure}[t]{0.54\textwidth}
    \centering
    \includegraphics[width=\textwidth]{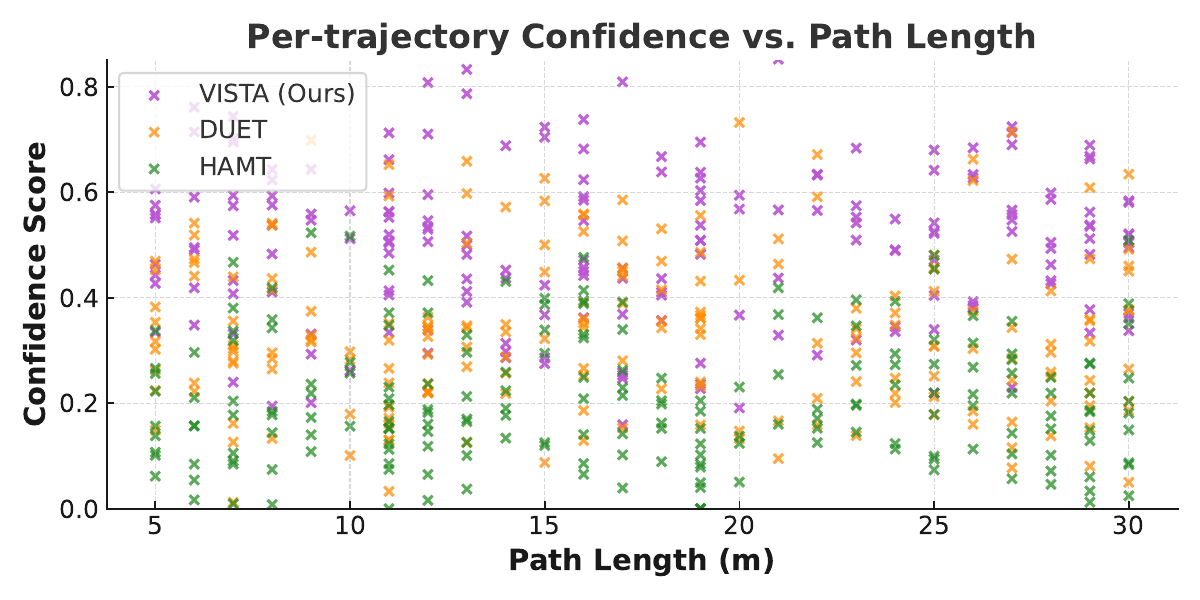}
    \caption{\textbf{Trajectory Confidence.} VISTA maintains more stable confidence across long paths.}
    \label{fig:confidence-length}
  \end{subfigure}
  \caption{\textbf{Ablation and Planning Analysis.} Bar chart and trajectory-level scatter plot illustrating the effect of removing key modules and VISTA’s robustness to path length.}
  \label{fig:ablation-figurepair}
\end{figure*}

 Table~\ref{tab:r2r_results} presents a comprehensive comparison on the Room-to-Room benchmark. VISTA consistently achieves state-of-the-art performance across both Validation and Test Unseen splits, outperforming prior models in SR, SPL, and NE.
On Validation Unseen, VISTA attains 77.8\% SR and 68.3 SPL, surpassing the previous best (DUET-Imagine) by \textbf{+3.2\% SR} and \textbf{+7.5 SPL}. On Test Unseen, VISTA maintains a strong 74.9\% SR, indicating that its improvements generalize robustly to held-out environments. These improvements stem from the structured integration of AIS, PAF, and CoT, enabling anticipatory and interpretable decision making. As shown in Figure \ref{fig:confidence-length}, VISTA maintains higher success rates, more efficient paths, and more stable confidence across increasing path lengths. This indicates that our model scales robustly to long-horizon tasks by combining global imagination, local grounding, and structured reasoning.
To further validate the generalization capability, we additionally evaluate VISTA on RoboTHOR \cite{deitke2020robothoropensimulationtorealembodied, samir2022cows} Table \ref{tab:robothor-table}. Compared to competitive baselines such as VLTNet and ESC, VISTA achieves significant improvements, reaching 28.8\% SWPL and 43.1\% SR. Qualitative comparisons Figure \ref{fig:robothor-image} demonstrate that our approach effectively grounds instructions into visual observations, highlighting its interpretability and robustness across different embodied navigation scenarios.

Despite significant performance improvements, VISTA occasionally encounters navigation failures. Typical failure scenarios include: 1. Inaccurate Visual Imagination: When the visual imagination module generates scenes significantly diverging from the actual environment, the subsequent perceptual alignment misleads action selection, causing navigation errors (e.g., selecting incorrect rooms or hallways). 2. Attention Misalignment: The perceptual alignment filter sometimes produces overly broad or misplaced attention masks, particularly in visually cluttered or ambiguous scenes, reducing localization accuracy and causing inefficient navigation paths. 3. Over reliance on Instruction-driven Mode: Occasionally, AIS prematurely selects static imagination mode in highly ambiguous environments, neglecting real-time visual cues and resulting in suboptimal paths or failed navigation.

\subsection{Ablation Studies}

We observe that removing the visual imagination module leads to the most severe performance drop, highlighting its role in enabling forward-looking behavior. Without predicted goal images, the agent struggles to reason beyond its current field of view, resulting in both lower success and increased 
\noindent
\begin{wraptable}{r}{0.4\textwidth}
  \vspace{-1em}
  \centering
  \caption{
    \small
    \textbf{Validation Unseen Ablations.} Removing imagination and alignment significantly reduces SR and SPL.
  }
  \label{tab:ablation-table}
  \vspace{0.3em}
  \small
  \begin{tabular}{lcccc}
    \toprule
    \textbf{Variant} & \textbf{SR} & \textbf{SPL} & \textbf{NE} & \textbf{TL} \\
    \midrule
    Full       & \textbf{77.8} & \textbf{68.3} & \textbf{2.92} & \textbf{13.26} \\
    w/o Img    & 66.7 & 61.0 & 3.98 & 12.55 \\
    w/o Filter & 68.3 & 62.1 & 3.42 & 14.30 \\
    w/o AIS    & 73.4 & 63.0 & 3.27 & 13.87 \\
    w/o CoT    & 71.5 & 64.4 & 3.18 & 13.76 \\
    \bottomrule
  \end{tabular}
  \vspace{-1em}
\end{wraptable}localization error. In contrast, removing the chain-of-thought (CoT) reasoning has a milder impact on success rate but noticeably reduces path efficiency. This suggests that while the agent can still reach the goal, it lacks the structured decision process to do so efficiently. Similarly, dropping the perceptual alignment filter degrades spatial focus, as the agent loses the ability to localize its imagined target within the scene—leading to more erratic trajectories. These trends confirm that each module in VISTA supports a distinct cognitive capacity, and removing any one of them disrupts the closed-loop interplay between imagination, grounding, and reasoning. To further understand module roles, we analyze their contributions via RoboTHOR, Visual imagination significantly improves SR (+11.1\%) on long paths ($>$15m), confirming its necessity for foresight in extended navigation.


\section{Conclusion}
\label{sec:conclusion}
We present \textbf{VISTA}, a closed-loop vision-and-language navigation framework that integrates visual imagination, perceptual grounding, and structured reasoning into a unified decision process. Central to our method is an adaptive imagination scheduler that dynamically selects between static and dynamic goal prediction based on navigation uncertainty and visual-semantic alignment. Imagined scenes are generated online using LoRA-finetuned Stable Diffusion, aligned with egocentric observations through a perceptual filter, and used to guide chain-of-thought reasoning in action selection. Our results on R2R and RoboTHORdemonstrate state-of-the-art performance, particularly in long-horizon and visually ambiguous scenarios. Through ablation and qualitative analysis, we show that each module—imagination, filtering, and reasoning—contributes distinctly to navigation accuracy, interpretability, and robustness.

Beyond VLN, our approach offers a general framework for visually grounded, imagination-driven planning. We believe that coupling generation and reasoning will be increasingly important for embodied agents operating under uncertainty, and we hope VISTA provides a step in that direction.

 \section{Limitations}

While VISTA shows strong performance in long-horizon navigation tasks, several limitations remain. First, the fidelity of imagined goal scenes depends on the generative quality of the diffusion model, which may fail in cluttered or diverse environments, leading to misalignments. Second, the imagination and reasoning pipeline introduces nontrivial computational overhead, which may limit real-time deployment. Third, the Adaptive Imagination Scheduler relies on manually tuned thresholds that may require adaptation across environments. Finally, all evaluations are conducted in simulation; real-world deployment may face additional challenges such as sensor noise and domain shifts. Future work may explore lightweight generation modules, learned scheduling strategies, and transfer methods to improve generalization and deployment readiness.


\clearpage


\bibliography{references}  

\end{document}